\RequirePackage{fix-cm}
\documentclass[smallextended]{svjour3}       % onecolumn (second format)
\smartqed  % flush right qed marks, e.g. at end of proof
\usepackage{graphicx}
\usepackage{amsmath}
\usepackage{amssymb}
\newcommand{\etal}{et al.}%

\def\bg{{\tt bg}}
\def\fg{{\tt f\!g}}
\def\fu{{\tt f\!u}}

\newcommand{\e}{\emph}
\newcommand{\tb}{\textbf}

\def\bx{\mathbf{x}}
\def\bc{\mathbf{c}}
\def\bb{\mathbf{b}}
\def\bf{\mathbf{f}}
\def\bo{\mathbf{0}}
\def\mNB{\mathcal{N_B}}
\def\mNF{\mathcal{N_F}}

\def\SigCB{\Sigma^{B}_{\mathbf C}}
\def\SigXB{\Sigma^{B}_{\mathbf S}}
\def\SigCF{\Sigma^{F}_{\mathbf C}}
\def\SigXF{\Sigma^{F}_{\mathbf S}}

\begin{document}

\title{Background subtraction - separating the modeling and the inference%\thanks{Grants or other notes
}


\author{Manjunath Narayana         \and
        Allen Hanson \and
        Erik G. Learned-Miller
}


\institute{Manjunath Narayana \at
%              University of Massachusetts Amherst\\
              \email{narayana@cs.umass.edu}           %  \\
           \and
           Erik G. Learned-Miller \at
%              University of Massachusetts Amherst\\
              \email{elm@cs.umass.edu}           %  \\
           \and
           Allen Hanson \at
              \email{hanson@cs.umass.edu}        \\   %  \\
              University of Massachusetts Amherst
}

\date{Received: date / Accepted: date}

\maketitle

\begin{abstract}
In its early implementations, background modeling was a process of 
building a model for the background of a video with a stationary camera,
and identifying pixels that did not conform well to this model. The 
pixels that were not well-described by the background model were assumed
to be moving objects.
Many systems today
maintain models for the foreground as well as the background, and these
models compete to explain the pixels in a video. If the foreground model
explains the pixels better, they are considered foreground. Otherwise they
are considered background.

In this paper, we argue that the logical endpoint of this evolution is to
simply use Bayes' rule to classify pixels. In particular, it is essential
to have a background likelihood, a foreground likelihood, and a prior at
each pixel. A simple application of Bayes' rule then gives 
%the minimum
%error criterion for the label, or, if desired, 
a posterior probability over the label. 
The only remaining question is the quality of the component 
models: the background likelihood, the foreground likelihood, and the prior.

We describe a model for the likelihoods that is built by using not
only the past observations at a given pixel location, but by also
including observations in a spatial neighborhood around the
location. This enables us to model the influence between neighboring
pixels and is an improvement over earlier pixelwise models that do
not allow for such influence.  Although similar in spirit to the joint
domain-range model, we show that our model overcomes certain
deficiencies in that model.  We use a spatially dependent prior for
the background and foreground. The background and foreground labels
from the previous frame, after spatial smoothing to account for
movement of objects, are used to build the prior for the current
frame. 

These components are, by themselves, not novel aspects in
background modeling.  As we will show, many existing systems account
for these aspects in different ways. We argue that separating these
components as suggested in this paper yields a very simple and
effective model. Our intuitive description also isolates the model
components from the classification or inference step. Improvements to
each model component can be carried out without any changes to the
inference or other components.  The various components can hence be
modeled effectively and their impact on the overall system understood
more easily.

\end{abstract}
%-------------------------------------------------------------------------
%-------------------------------------------------------------------------
\section{Introduction}\label{sec:intro}
Background subtraction for stationary camera videos is a well
researched problem. Algorithms have evolved from early approaches
modeling the background at each
pixel~\cite{Wren97,Toyama99,Stauffer99,Elgammal00} to methods
that include an explicit model for the foreground~\cite{Li03,Sheikh05},
and finally to more recent models that incorporate spatial dependence
between neighboring pixels~\cite{Sheikh05}.

In early algorithms~\cite{Wren97,Stauffer99}, a probability distribution 
$p_\bx(\bc|\bg)$ over {\em background} colors $\bc$ is defined and 
learned for each location $\bx$ in the image. These distributions are 
essentially the background likelihood at each pixel location.
Pixels that are well explained by
the background likelihood are classified as \emph{background} and the
remaining pixels in the image are labeled as \emph{foreground}.
Toyama~\etal~\cite{Toyama99} use a Weiner filter to predict the
intensities of the background pixels in the current frame using the
observed values from the previous frames and to identify
non-conforming pixels as foreground.  
Wren \etal~\cite{Wren97} model the background
as a Gaussian distribution at each pixel.  To account for the multiple
intensities often displayed by background phenomena such as leaves
waving in the wind or waves on water surfaces, Stauffer and
Grimson~\cite{Stauffer99} learn a parametric mixture of Gaussians
(MoG) model at each pixel.  The MoG model update procedure as
described by Stauffer and Grimson can be unreliable during initialization
when not enough data have been observed.
To improve the performance during model initilization,
Kaewtrakulpong and Bowden~\cite{Kaewtrakulpong01} suggest a
slightly different model update procedure.  Porikli and
Tuzel~\cite{Porikli05} obtain the background likelihood by using a 
Bayesian approach to model the mean and variance values of 
the Gaussian mixtures.
Elgammal \etal~\cite{Elgammal00} avoid the drawbacks of using a parametric
MoG model by instead building the background likelihoods with 
non-parametric kernel density estimation (KDE) using data samples 
from previous frames in history.

While they are still called ``backgrounding'' systems, later systems
maintain a model for the foreground as well as the
background~\cite{Li03,Sheikh05}.  Explicit modeling of the foreground
has been shown to improve the accuracy of background
subtraction~\cite{Sheikh05}.  In these models, pixel labeling is
performed in a competitive manner by labeling as foreground the pixels
that are better explained by the foreground model. The remaining
pixels are labeled as background.

Although it is natural to think about priors along with likelihoods,
the use of an explicit prior for the background and foreground is less 
common. In the object tracking literature, Aeschliman \etal~\cite{Aeschliman10} 
use priors for the background and foreground objects for segmentation of 
tracked objects. 
In background modeling algorithms that do not explicitly model the prior,
the foreground-background likelihood ratio is used for classification. 
Pixels that have a likelihood ratio greater than some
predefined threshold value are labeled as foreground. This method is 
equivalent to using an implicit prior that is the same at all pixel 
locations. 

Thus, existing algorithms make use of some subset of the three natural 
components for background modeling - the background likelihood, 
the foreground likelihood, and the prior. They make up for the missing
components by including effective model-specific procedures at the
classification stage.
For instance, Elgammal \etal~\cite{Elgammal00}\ and Stauffer and 
Grimson~\cite{Stauffer99} use only the background likelihood, but, 
during classification, consider a likelihood threshold below which 
pixels are considered as foreground. 
Zivkovic~\cite{Zivkovic04} describes Bayes' rule for computing
background posteriors, but since neither the foreground likelihood nor
the priors are explicitly modeled, the classification is essentially
based on a threshold on background likelihood values.
Sheikh and Shah~\cite{Sheikh05} utilize both foreground and background 
likelihoods, but do not use an explicit prior. 
Instead, by using a foreground-background 
likelihood ratio as the classification criterion, they effectively use a 
uniform prior.

We argue that the logical endpoint of the model evolution for
backgrounding is a system where all three components are explicitly
modeled and Bayes' rule is applied for classification.  Such a system
has the advantage of being a simpler model where the modeling of the
individual components is isolated from the inference step.  This
separation allows us to describe the components without any relation
to the classification procedure. Our motivation behind this approach
is that the components can individually be improved, as we will show
in later sections, without affecting each other or the final
inference procedure.

In the rest of the paper, we describe the components of our background system 
and place them in the context of existing algorithms where possible. Section 
\ref{sec:bgLik} discusses the evolution of the background likelihood models 
and our improvements to the most successful models. 
In Section \ref{sec:fgLik}, we discuss the modifications to the likelihood 
for modeling the foreground. 
Modeling of the prior is described in Section \ref{sec:prior}.
Computation of posterior probabilities by using the above components is 
explained in Section \ref{sec:bayes}.
%Our recent improvements to the likelihood model by using an adaptive 
%variance at each pixel location is presented in Section \ref{sec:adaptive}.
Results comparing our system to earlier methods on a benchmark data set 
are given in Section \ref{sec:earlyResults}. 
Recent improvements to the background likelihood and its impact on the 
system's accuracy are described in Sections \ref{sec:adaptive} and
\ref{sec:results}. 
We conclude with a discussion in Section \ref{sec:discussion}.
%-------------------------------------------------------------------------
%-------------------------------------------------------------------------
\section{Background likelihood}\label{sec:bgLik}
The background likelihood, which is a distribution over feature values,
is a common aspect in many backgrounding systems.  Stauffer and Grimson 
~\cite{Stauffer99} model the
background likelihood at each pixel using a mixture of Gaussians (MoG)
approach. 
The requirement of specifying the number of mixture components in
the MoG model is removed in the non-parametric kernel density estimation
(KDE) model~\cite{Elgammal00}.  
In the KDE model, the distributions at each pixel
location are estimated by summing up contributions from the observed
background data samples at that location from previous frames in
history.  
For each pixel location $\bx=[x,y]$, both these models maintain a
distribution $p_\bx(\bc)$ that is independent of the neighboring pixels. 
Here, $\bc=[r, g, b]$ is a vector that represents color.
These neighbor-independent distributions have the drawback of not being 
able to account for the influence of neighboring pixels on each other's
color distributions.

To allow neighboring pixels to influence the background likelihood at a 
given pixel location, we model the likelihood at a particular pixel 
location to be a weighted sum of distributions from its spatial neighbors.
Our \e{smoothed} background likelihood $P_\bx(\bc)$ for each pixel 
location $\bx$ is a weighted sum of distributions from a spatial 
neighborhood $\mNB$ around $\bx$. 
Each neighboring likelihood is weighted by its spatial distance
(i.e., distance in the image coordinates) from $\bx$:
\begin{equation}\label{eq:P_x_smoothed} 
P_\bx(\bc|\bg;\SigXB) = \frac{1}{Z}
       \sum_{\Delta\in \mNB} p_{\bx+\Delta}(\bc|\bg)
          \times G(\Delta; \bo, \SigXB).
\end{equation}
Here $\Delta$ is a spatial
displacement that defines a spatial neighborhood $\mNB$ around the pixel 
location $\bx$ at which the likelihood is being computed. 
$G(\cdot; \bo,\SigXB)$ is a zero-mean multivariate Gaussian with 
covariance $\SigXB$. $B$ indicates that the covariance 
is for the background model and $\mathbf{S}$ denotes the spatial 
dimensions.
The normalization constant $Z$ is 
\begin{equation}\label{eq:Z_norm} 
Z = \sum_{\Delta\in \mNB} G(\Delta; \bo, \SigXB).
\end{equation}

\begin{figure}[t]
\begin{center}
   \includegraphics[width=1.0\linewidth]{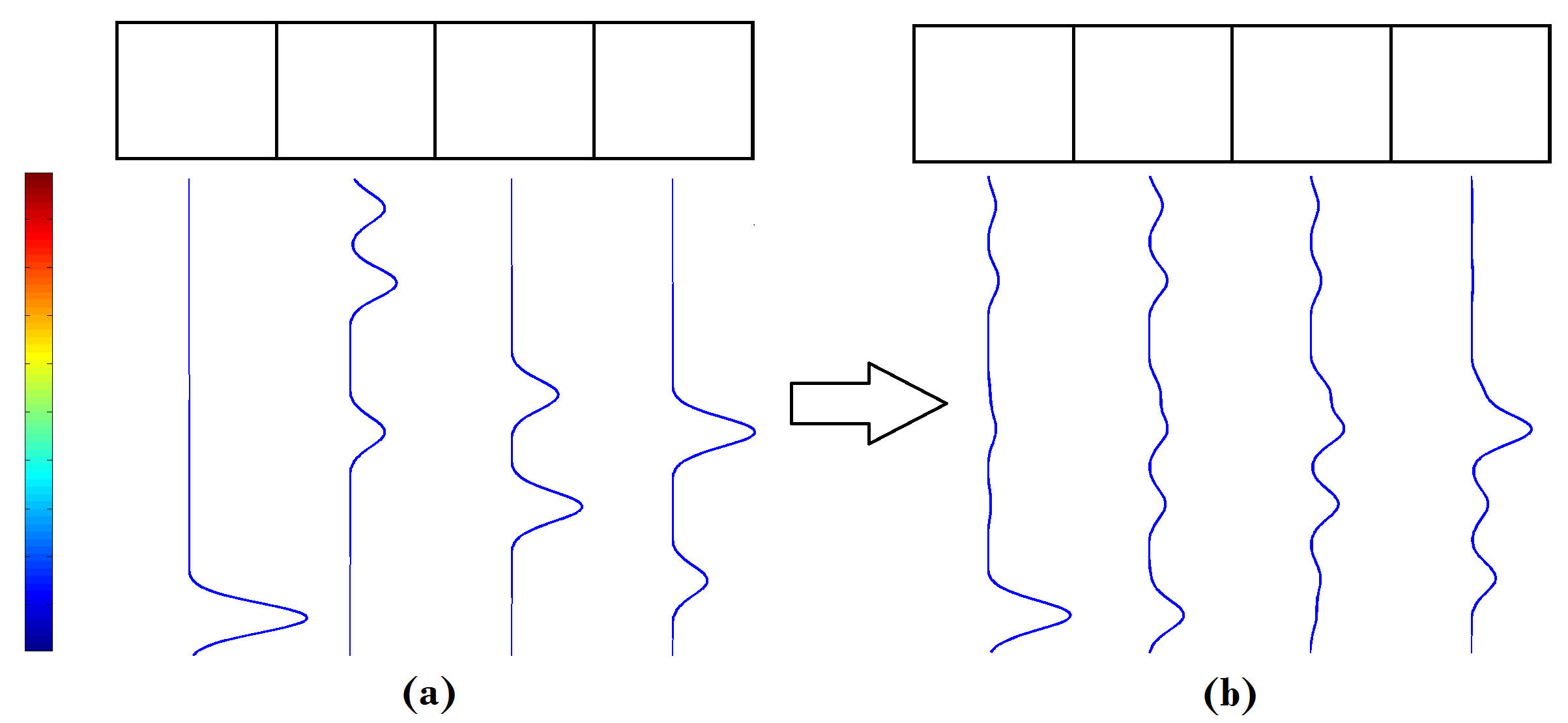}
\end{center}
   \caption{Influence of neighboring pixels on each other is modeled by
       spreading information spatially. 
       Figure (a) shows some example likelihoods for each pixel in a 
       single-dimensional (row) image.
       The distributions shown below each pixel are the estimated 
       background likelihoods. 
       The vertical axis corresponds to color values which are 
       visualized in the color map on the left side
       of the image. The horizontal axis corresponds to the probability 
       of the corresponding color.
       Figure (b) shows the smoothed likelihood at each pixel, which is 
       a weighted sum of the likelihoods in the pixel's neighborhood. 
       The effect of smoothing is clearly visible in the first pixel. 
       The distribution 
       in the first pixel clearly influences the distributions at the second
       and third pixels. The distance-dependent nature of the weights results
       in the first pixel influencing the third pixel less than it does the 
       second pixel.
   }
\label{fig:spatialSmoothing}
\end{figure}

The weighted sum results in a spatial smoothing of the distributions as
shown in Figure \ref{fig:spatialSmoothing}. This spreading of information
is useful in modeling spatial uncertainty of background pixels.
$\SigXB$ controls the amount of smoothing and spreading of information
in the spatial dimensions.

\begin{figure}[t]
\begin{center}
   \includegraphics[width=1.0\linewidth]{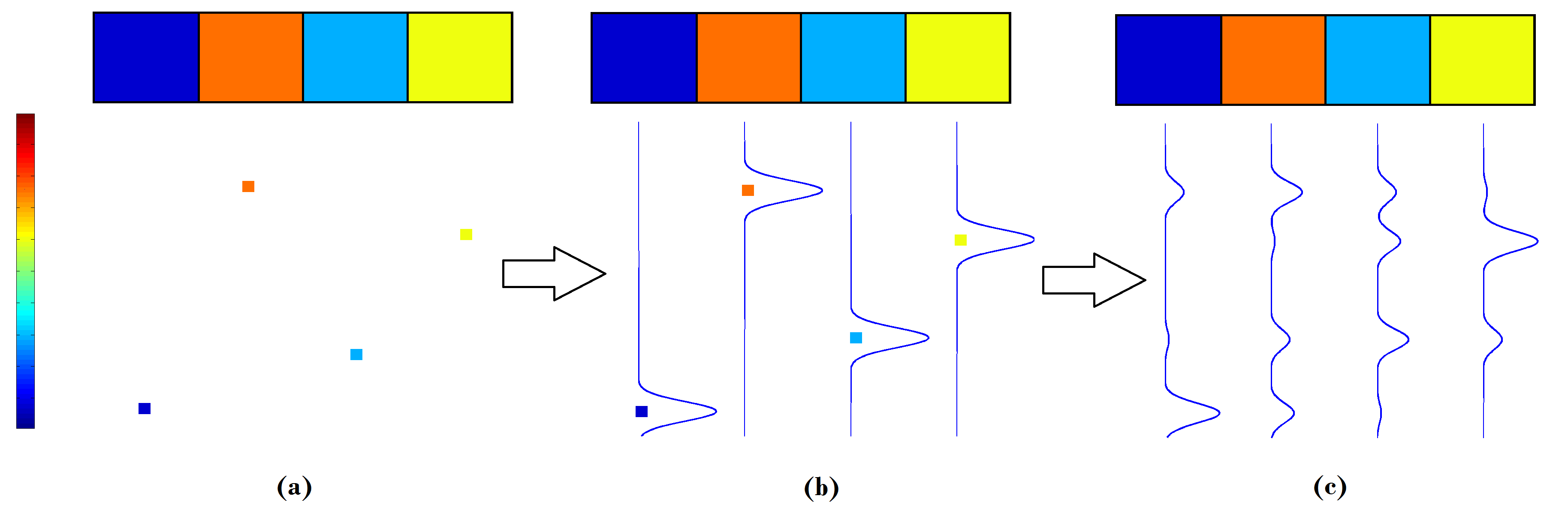}
\end{center}
   \caption{Modeling the likelihoods using pixel data samples and 
       KDE.
       Figure (a) shows the colors at each pixel. The corresponding 
       color and its location with respect to the vertical color axis 
       is shown under each pixel.
       Figure (b) shows the likelihood at each pixel estimated using 
       KDE with a Gaussian kernel.
       Figure (c) shows the effect of spatial smoothing of the KDE-based 
       likelihoods. Again, the illustration uses a one-dimensional row
       image in which a pixel's color is also represented in one dimension.
       It is straightforward to extend the example to two-dimensional 
       spatial coordinates and three-dimensional color space.
   }
\label{fig:KDEspatialSmoothing}
\end{figure}

Explicitly maintaining a distribution at each pixel location is impractical
for color features which can take one of $256^3$ values if each of the three 
color channels have a range between $0$ and $255$.
Instead, we compute likelihoods with KDE using the data samples from 
the previous frame. Let $\bb^{t-1}_{\bx}$ be the observed background color at 
pixel location $\bx$ in the previous frame.
Using a Gaussian kernel with covariance $\SigCB$ in the color dimensions,
our KDE background likelihood in the video frame numbered $t$ is given by
\begin{equation}\label{eq:KDE_P_BG} 
P^t_\bx(\bc|\bg;\SigCB,\SigXB) = \frac{1}{Z}
       \sum_{\Delta\in \mNB} G(\bc-\bb^{t-1}_{\bx+\Delta}; \bo, \SigCB)
          \times G(\Delta; \bo, \SigXB).
\end{equation}
Figure \ref{fig:KDEspatialSmoothing} illustrates the process of computing
the background likelihood using the observed background colors in one image.
It may be noted that the covariance matrix $\SigXB$ controls the amount 
of spatial influence from neighboring pixels. The covariance matrix
$\SigCB$ controls the amount of variation allowed in the color values of
the background pixels.

Finally, we consider background data samples not just from the previous
frame, but from the previous $T$ frames in order to obtain a more accurate
likelihood. We also allow probabilistic contribution from the previous 
frames' pixels by weighting each pixel according to its probability of 
belonging to the background:
\begin{equation}\label{eq:DF_KDE_bg} 
P^t_\bx(\bc|\bg;\!\Sigma^B)\!\! =\!\! \frac{1}{K_\bg}\!\!
       \sum_{i\in 1:T}\! \sum_{\Delta\in \mNB}\!\!\!\!\! G(\bc-\bb^{t-i}_{\bx+\Delta}; \bo,\! \SigCB)
       \times G(\Delta; \bo,\! \SigXB) \times P^{t-i}_\bx(\bg|\bb^{t-i}_{\bx+\Delta}).
\end{equation}

$\Sigma^B$ represents the covariance matrices for the background model
and consists of the color dimensions covariance matrix 
$\SigCB$ and the spatial dimensions covariance matrix $\SigXB$.
$P^t_\bx(\bg|\bb^{t}_{\bx})$ is the probability that pixel at location 
$\bx$ in the frame $t$ is background.
$K_{\bg}$ is the appropriate normalization factor:
\begin{equation}\label{eq:bg_norm}
K_{bg} = \sum_{i\in 1:T} \sum_{\Delta\in \mNB} G(\Delta; \bo, \SigXB) \times
    P^{t-i}_\bx(\bg|\bb^{t-i}_{\bx+\Delta}).
\end{equation}
For efficiency, we restrict the covariance matrices to be diagonal and
hence parameterize them by their diagonal elements.

\subsection{Existing work on spatial smoothing of distributions}
The use of spatial smoothing of distributions is not entirely new. 
Sheikh and Shah~\cite{Sheikh05} use a 
joint domain-range model that combines the pixels' position values and 
color observations into a joint five-dimensional space.
By modeling the likelihoods in the joint space, they
allow pixels in one location to influence the
distribution in another location.
Their background likelihood is defined as:\footnote{We have modified their
equation to allow probabilistic contributions from the pixels and changed
the notation to make it easily comparable to ours.}
\begin{equation}\label{eq:SS_KDE_bg} 
\begin{split}
P^t(\bc, \bx|\bg;\Sigma^B)\! =\! \frac{1}{K}\!\!
    \sum_{i\in1:T}\sum_{\Delta\in \mNB}
    \!\!\!\!G(\bc-\bb^{t-i}_{\bx+\Delta};\bo, \SigCB) G(\Delta;\bo, \SigXB) 
    P^{t-i}_\bx(\bg|\bb^{t-i}_{\bx+\Delta}).
\end{split}
\end{equation}
The normalization constant, $K$, is given by
\begin{equation}\label{eq:SS_norm} 
K = \sum_{i\in1:T}\sum_{\Delta\in \mNB}
    P^{t-i}_\bx(\bg|\bb^{t-i}_{\bx+\Delta}).
\end{equation}

%\begin{equation}\label{eq:SS_KDE_bg} 
%\begin{split}
%&P^t(\bc, \bx|\bg;\Sigma^B) =\\
%&\frac{1}{\sum_{i\in1:T}\sum_{\Delta\in \mNB}
%    P^{t-i}_\bx(\bg|\bb^{t-i}_{\bx+\Delta})}
%    \sum_{i\in1:T}\sum_{\Delta\in \mNB}
%    G(\bc-\bb^{t-i}_{\bx+\Delta};\bo, \SigCB) G(\Delta;\bo, \SigXB) 
%    P^{t-i}_\bx(\bg|\bb^{t-i}_{\bx+\Delta}).
%\end{split}
%\end{equation}

The key difference between their model and ours is that theirs is, for the 
entire image, a \emph{single} distribution in the joint domain-range space 
whereas ours consists of a different location-dependent distribution at each pixel.
This difference has a big effect on the classification stage.
As we will see later, their classification criterion, based on the ratio of
foreground and background likelihoods in this five-dimensional space, 
has an undesirable dependence on the size of the image. 
By replacing the single joint distribution with a \emph{field} of 
distributions dependent on image location, we avoid the 
dependence on image size and achieve better results.

The joint domain-range model has been used earlier in the object tracking 
literature. Elgammal \etal~\cite{Elgammal03} use a 
joint domain-range model that is almost identical to the background 
model of Sheikh and Shah~\cite{Sheikh05}.
A scheme very similar to our Equation \ref{eq:P_x_smoothed} was used in 
a tracking system by Han and Davis~\cite{Han05} to interpolate the 
pixelwise appearance distributions for an object whose size has changed 
during the tracking process.
The close resemblance between these models suggests that tracking and 
background modeling share similar fundamental principles and can be
achieved under the same framework. One such framework that integrates
segmentation and tracking has been described by 
Aeschliman \etal~\cite{Aeschliman10}.

Ko \etal~\cite{Ko08} use a histogram-based variant of the Sheikh and Shah
~\cite{Sheikh05} background model which is built from observations 
in a spatial neighborhood around each pixel from previous frames 
in history. 
However, they do not consider the spatial distance between a pixel and its
neighbor when summing up the contributions. 
In addition, they build another distribution, which can be interpreted as 
the ``texture'' at each pixel, by using only the current frame 
observations in each pixel's spatial neighborhood.
Their classification criterion for foreground pixels is to threshold the
Bhattacharya distance between the background 
distribution and the ``texture'' distribution. 
Our model is different because of our classification criterion that uses
foreground likelihoods and explicit priors for the background and foreground
which we discuss in subsequent sections.
%Sevilla and Learned-Miller~\cite{Sevilla12} use the term ``distribution 
%field'' to describe models that consist of a distribution at each pixel 
%and where the distributions are estimated from local neighborhoods.
%We henceforth refer to our likelihood model as a distribution field 
%likelihood model.
%-------------------------------------------------------------------------
%-------------------------------------------------------------------------
\section{Foreground likelihood}\label{sec:fgLik}
Explicit modeling of the foreground likelihood has been shown to result in
more accurate systems~\cite{Li03,Sheikh05}. Our foreground likelihood is 
very similar to our background likelihood. However, it is important to 
consider in the foreground likelihood, the possibility of hitherto unseen color
values appearing as foreground. This may happen because a new foreground
object enters the scene or an existing foreground object either changes
color or, by moving, exposes a previously unseen part of it. 
We find it useful to separate the foreground process into two different 
sub-processes: previously seen foreground, which we shall refer to as
\emph{seen} foreground, and previously unseen foreground, which we shall
refer to as \emph{unseen} foreground. 
The likelihood for the seen foreground process is computed using a KDE
procedure similar to the background likelihood estimation:
\begin{equation}\label{eq:DF_KDE_fg} 
P^t_\bx(\bc|\fg;\!\Sigma^F)\! =\! \frac{1}{K_\fg}\!
       \sum_{i\in 1:T}\! \sum_{\Delta\in \mNF}\!\!\!\! G(\bc-\bf^{t-i}_{\bx+\Delta}; \bo,\! \SigCF)
       \times G(\Delta; \bo,\! \SigXF) \times P^{t-i}_\bx(\fg|\bf^{t-i}_{\bx+\Delta}).
\end{equation}
Similar to Equation \ref{eq:DF_KDE_bg}, $\bf^{t}_{\bx}$ is the observed 
foreground color at pixel location $\bx$ in frame $t$.
$\Sigma^F$ is the covariance matrix for the foreground
model, and $K_{\fg}$ is the normalization factor, analogous to $K_{\bg}$.
$P^t_\bx(\fg|\bf^{t}_{\bx})$ is the probability that pixel at location 
$\bx$ in the frame $t$ is foreground.

Since foreground objects typically move more than background objects and 
also exhibit more variation in their color appearance, we typically 
use higher covariance values for the foreground than for the background.

The likelihood for the unseen foreground process is simply a uniform 
distribution over the color space.
\begin{equation}\label{eq:DF_KDE_Ufg} 
P^t_\bx(\bc|\fu) = \frac{1}{R\times G\times B}
\end{equation}
for all locations $\bx$ in the image, where $R$, $G$, and $B$,
are the number of possible intensity values for red, green, and blue colors 
respectively.

The unseen foreground process constantly tries to account as foreground 
any colors not reasonably explained by both the background and the seen 
foreground likelihood.

The concept of using a uniform likelihood is not new. For instance, Sheikh
and Shah~\cite{Sheikh05} mix a uniform distribution 
(in five-dimensional space) to their foreground likelihoods to explain 
the appearance of new foreground colors in the
scene. Separation of the foreground process into two sub-processes, as we 
have done, is equivalent to the mixing of the likelihoods into one 
combined likelihood. The advantage of considering them as separate 
sub-processes is that when combined with a separate prior for each, 
greater modeling flexibility can be achieved. For instance, at image 
boundaries where new objects tend to enter the scene, a higher
prior can be used for the unseen foreground process.
%-------------------------------------------------------------------------
%-------------------------------------------------------------------------
\section{Priors}\label{sec:prior}
In addition to modeling the likelihoods, we explicitly model spatially
varying priors for the background and foreground processes. Such spatial 
priors have recently been used for segmentation of objects being followed
in a tracking algorithm~\cite{Aeschliman10}. 
Background modeling systems that use a likelihood ratio as the 
classification criterion are implicitly assuming a uniform prior for the
entire image.
In such systems, if the foreground-background likelihood ratio at a given pixel
is greater than some predefined threshold $L$, then the pixel is labeled
as foreground. Using a value of $1$ for $L$ means that the background and 
foreground processes have a uniform and equal prior value at every pixel 
location. Other values of $L$ imply using a uniform
but unequal prior for the background and foreground.

We generalize the notion of the prior by considering a spatially varying
prior. 
The uniform prior is simply a special case of our model.
We define pixelwise priors for the three processes involved 
- background, previously seen foreground, and unseen foreground. 
The classified pixel 
labels from the previous frame are used as a starting point for building 
the priors for the current frame.
We assume that a pixel that is classified as background in the
previous frame has a 95\% probability of belonging to the background in the current
frame as well. The pixel has a 2.5\% probability of belonging to a seen
foreground object, and a 2.5\% probability of coming from a previously
unseen foreground object. 
For a foreground pixel in the previous frame, we assume that
due to object motion, there is a 50\% probability of this pixel
becoming background, a 25\% probability of this pixel belonging to the same
foreground object as in the previous frame, and a 25\% probability that it
becomes a new unseen object.

\begin{figure}[t]
\begin{center}
   \includegraphics[width=1.0\linewidth]{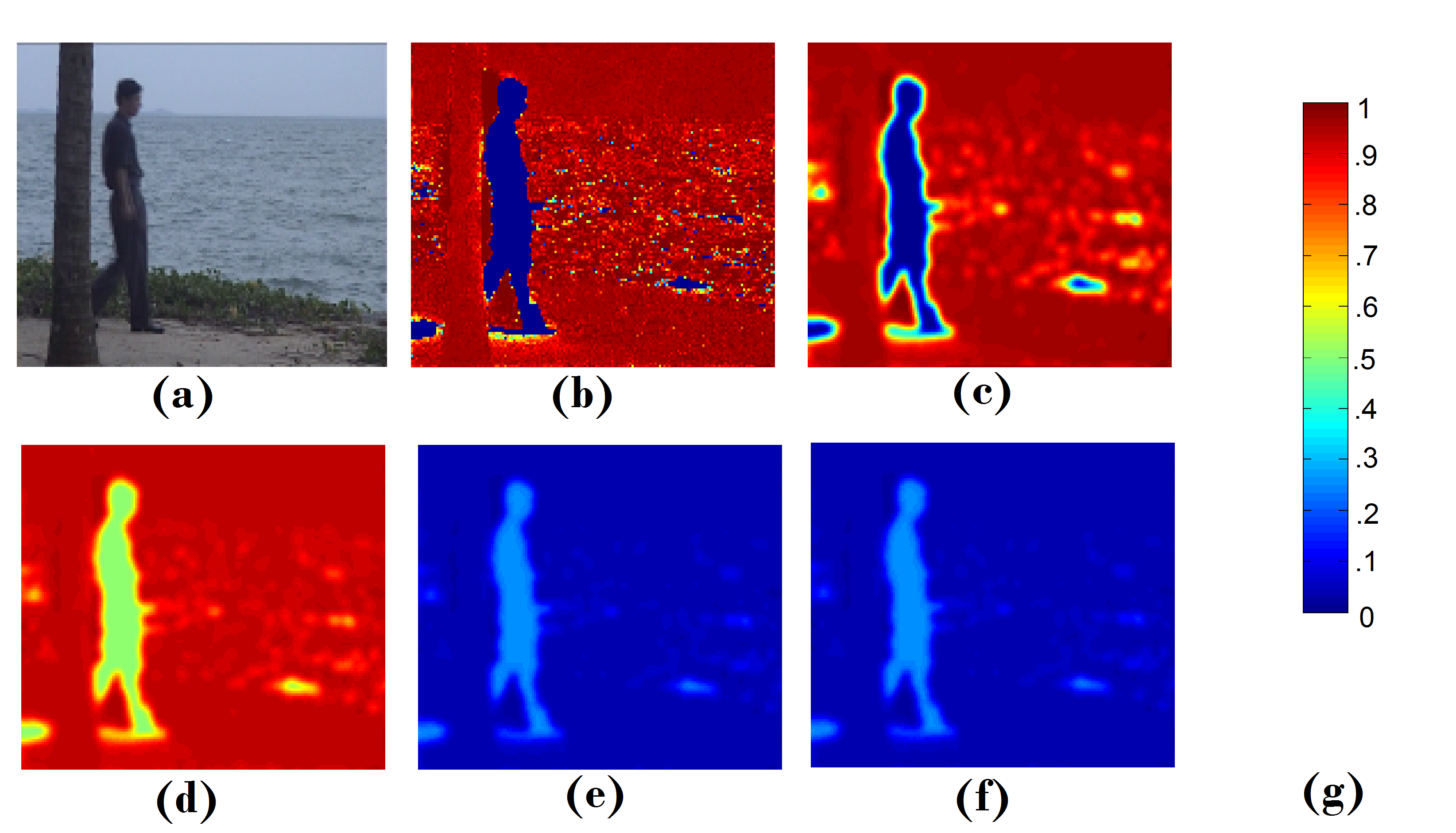}
\end{center}
   \caption{Illustration of computation of the spatially dependent prior.
       The image from the previous frame is shown in (a).
       The background probabilities in (b) are first smoothed with a 
       Gaussian filter to allow for some amount of object motion in 
       the scene.
       The smoothed probabilities are shown in (c), from which 
       the background prior (d), the foreground prior (e), and the 
       unseen foreground prior (f) are computed. 
       The mapping from color to probability values is given in (g).
       We use equivalent equations for the foreground and unseen foreground
       priors which result in (e) and (f) being identical.
   }
\label{fig:prior_ex}
\end{figure}

There are hence essentially two settings for the prior at each pixel 
depending on whether the pixel was labeled background or foreground in the
previous frame. Instead of using the hard thresholds described above,
we use the pixel's background label probability from the previous frame
when computing the prior.
For instance, a pixel that has probability $p$ of being background
in the previous frame will have a background prior equal to $p\times.95 +
(1-p)\times .5$. 
Also, since objects typically move by a few pixels 
from the previous frame to the current frame, we apply a smoothing
($7\times 7$ Gaussian filter with a standard deviation value of $1.75$) 
to the classification results from the previous frame
before computing the priors for the current frame. Let
$\tilde{P}^{t-1}_\bx(\bg)$ be the smoothed background posterior image from the
previous frame. The priors for the current frame are
\begin{equation}\label{eq:priors}
\begin{split}
P^{t}_\bx(\bg) = &\tilde{P}^{t-1}_\bx(\bg)\times.950 +
(1-\tilde{P}^{t-1}_\bx(\bg))\times.500,\\
P^{t}_\bx(\fg) = &\tilde{P}^{t-1}_\bx(\bg)\times.025 +
(1-\tilde{P}^{t-1}_\bx(\bg))\times.250,\\
P^{t}_\bx(\fu) = &\tilde{P}^{t-1}_\bx(\bg)\times.025 +
(1-\tilde{P}^{t-1}_\bx(\bg))\times.250.\\
\end{split}
\end{equation}
Figure \ref{fig:prior_ex} is an illustration of the prior computation 
process. Figure \ref{fig:prior_ex}(a) shows the previous frame for 
which the background label probabilities at each pixel have been computed 
in (b).
The background probabilities are smoothed with a Gaussian filter in (c).
Using Equation \ref{eq:priors}, the background prior (d), the 
foreground prior (e), the unseen foreground prior (f) are computed.
These priors are then used for computing the posterior probabilities
in the current frame, as we explain in the next section.

In  our implementation, although the likelihoods for the
foreground and unseen foreground processes are different, the priors for 
the two processes are equal at every pixel.
It is not necessary that the priors for the seen foreground and the 
unseen foreground be the same in all background modeling systems.  
For instance, at image boundaries, using a higher prior value for the unseen
foreground could result in better detection of new objects that enter
the scene in these regions.

Our choice of the values $.95$ and $.50$ for the background prior for 
pixels that have been labeled as background and foreground in the previous
frame respectively is guided by the intuition that background pixels change
their label from one frame to the next very rarely and foreground objects 
that are moving have a moderate chance of revealing the background in the 
next frame. That these values are set by hand is a weakness of our current
system.\footnote{Observations from the ground truth labels from videos
in the change detection data set~\cite{Goyette12} show that between 95 
and 100 percent of all pixels labeled as background in each frame retain 
their background label in the next frame. We believe the use of the value
$.95$ for background prior is justified in light of this observation.
The use of $.50$ for the background prior in pixel locations that were
labeled as foreground in the previous frame essentially allows the 
likelihood to decide the labels of these pixels in the current frame.}
The advantage of our approach is that these values can easily be
learned automatically by accumulating statistics from the scene over a 
long period of time. Although the effect of using different priors for
the background and foreground is equivalent to using a decision 
threshold on the foreground-background likelihood ratio, the priors are
easier to understand and update. 
For example, the priors at each pixel can be updated using the  
statistics of pixel labels from long term scene history. 
The statistics could reveal a 
higher foreground prior near doors in the scene and at image borders.
A similar scheme to update a decision threshold at 
these locations is far less natural.

We use a Gaussian filter of size $7$ because the foreground objects in 
these videos typically move by $5$ to $10$ pixels. The size of the filter 
can potentially be learned by tracking the foreground objects. 
If there is a significant depth variation in different parts of the scene, 
a different parameter can be learned for the corresponding image regions by 
using tracking information~\cite{Narayana07}.

\section{Computing the posteriors - putting the components together during 
    inference}\label{sec:bayes}

Given the likelihoods and the priors as described in the previous sections,
the only thing left to do is to compute the posterior probability of
background and foreground, conditioned on the observed pixel values 
using Bayes' rule.

Given an observed color vector $\bc$ at pixel location $\bx$ in frame $t$, 
the probability of background and foreground are
\begin{equation}\label{eq:bayes}
\begin{split}
&P^t_\bx(\bg|\bc) = \frac{P^t_\bx(\bc|\bg; \Sigma^B)\times P^t_\bx(\bg)}
{\sum_{l=\bg, \fg} P^t_\bx(\bc|l; \Sigma^l)\times P^t_\bx(l) 
%+P(a_{c}|f\!g, a_x, a_y; \Sigma^F)\times P(f\!g|a_x, a_y)
+P^t_\bx(\bc|\fu)\times P^t_\bx(\fu)}\\
&P^t_\bx(\fg) = 1-P^t_\bx(\bg|\bc).
\end{split}
\end{equation}

When the ideal likelihoods and priors are known, classification based on
Bayes' rule gives the minimum possible error. 
A common alternative classification criterion is the ratio of
the foreground likelihood to the background likelihood.
The likelihood ratio classification in the joint 
domain-range model deserves special consideration because it implicitly includes 
a notion of a prior. 
However, as we show in the next section, the implicit prior involved 
causes a peculiar dependence on the image size. Our model does not exhibit this
undesired consequence.
\subsection{Likelihood ratio-based classification in the joint domain-range model}\label{subsec:domRangeClassifier}
In the Sheikh and Shah joint domain-range model~\cite{Sheikh05}, 
the classification of pixels 
is done based on the likelihood ratios of the background and foreground processes. 
The decision criterion based on the ratios of the five-dimensional
background and foreground likelihoods can be represented as 
\begin{equation}\label{eq:Sheikh_ratio}
\begin{split}
P^t(\bc, \bx|\bg) &\overset{?}{\gtrless} P^t(\bc, \bx|\fg)\\
P^t(\bc|\bx,\bg)\times P^t(\bx|\bg) &\overset{?}{\gtrless} P^t(\bc|\bx,\fg)\times
P^t(\bx|\fg).
\end{split}
\end{equation}
The classification decision hence depends on the factors $P^t(\bx|\bg)$
and $P^t(\bx|\fg)$. 
These factors are the prior probability of a particular pixel location given
the background or foreground process. 
For any pixel location $\bx$, these
factors can depend upon parts of the image that are arbitrarily far away.
This is because the prior likelihood of a given pixel location being 
foreground will be smaller if more pixels from another part of the image 
are detected as foreground, and larger if fewer pixels elsewhere are detected 
as foreground (since $P^t(\bx|\fg)$ must integrate to 1). 
Furthermore, these factors will change when the image size is changed, hence
affecting the classification~\cite{Narayana12BMVC}. 
By separating the system components and bringing them together 
during the posterior computation, we avoid this arbitrary dependence 
on the size of the image.
%-------------------------------------------------------------------------
%-------------------------------------------------------------------------
\section{Comparison to earlier systems}\label{sec:earlyResults}
In this section, we compare our system to the various earlier
systems described in the paper so far. We use the I2R benchmark data set~\cite{Li03} 
with nine videos taken in different settings. The videos
have several challenging features like moving leaves and waves, strong
object shadows, and moving objects becoming stationary for a long
duration. 
The videos are between 500 and 3000 frames in length and typically
128 x 160 pixels in size.
Each video has $20$ frames for which the ground truth has been
marked. We use the F-measure to judge accuracy~\cite{Li10}; the higher the 
F-measure, the better the system:
\begin{equation}\label{eq:fmeasure}
F = \frac{2\times recall \times precision}{recall+precision}.\end{equation}
We use a Markov random field (MRF) to post-process the labels as is done
by Sheikh and Shah.
Further, to be consistent with the experimental set up of earlier 
systems~\cite{Li10,Narayana12},
we discard any foreground detections smaller than 15 pixels in size.
The various systems compared are the MoG model of Stauffer and 
Grimson~\cite{Stauffer99},
the KDE model of Elgammal \etal~\cite{Elgammal00}, the complex background-foreground model of
Li \etal(ACMMM03)~\cite{Li03}, the joint domain-range model of Sheikh and 
Shah (jKDE)~\cite{Sheikh05},
\footnote{The KDE and jKDE models are our own implementations and 
    include spatially-dependent priors and Bayes' classification 
    criterion in order to make a fair comparison.}
and our model, which we call the distribution field 
background (DFB) model.
The naming reflects the fact that our model is a \e{field} of 
distributions with one distribution at each pixel location and was 
inspired by the description of such models in the tracking literature 
by Sevilla-Lara and Learned-Miller~\cite{Sevilla12}.

Results in Table \ref{tbl:early_methods} show that systems that model 
the spatial influence of pixels, namely the jKDE  model and our 
DFB model yield significantly higher accuracy. 
The table shows that the jKDE system is most
accurate for our chosen parameter setting. Although this is not true for
other parameter settings,
\footnote{For a detailed comparison of our model 
and the joint domain-range model, the reader is referred to our earlier 
paper~\cite{Narayana12BMVC}.}
the table makes an 
important point that very effective systems can be built even if the 
underlying model has certain deficiencies (as we showed in Section
\ref{subsec:domRangeClassifier} for the jKDE). 
Mere separation of the model components as we have done and 
computing posterior probabilities for the labels does not 
guarantee better results. 
The usefulness of our system description is in the clear understanding of 
the different components and allowing for better modeling of the 
components without having to tweak the inference procedure. 
To illustrate this aspect of our system, we 
next describe one specific example of improving the background likelihood
model by identifying a shortcoming in the model and developing a strategy 
to fix it.
\begin{table}[ht]
\begin{center}
\begin{tabular}{|l|c|c|c|c|c|c|c|}
\hline
Video & MoG & KDE &ACMMM03 & jKDE  & DFB\\ 
      &     &     &        &       &    \\ 
\hline\hline
AirportHall     & $57.86$ & $62.46$ & $50.18$ & $70.13$ & $67.95$\\
Bootstrap       & $54.07$ & $61.15$ & $60.46$ & $71.77$ & $69.17$\\ 
Curtain         & $50.53$ & $61.83$ & $56.08$ & $87.34$ & $85.66$\\ 
Escalator       & $36.64$ & $40.84$ & $32.95$ & $53.70$ & $54.01$\\ 
Fountain        & $77.85$ & $52.76$ & $56.49$ & $57.35$ & $77.11$\\ 
ShoppingMall    & $66.95$ & $63.05$ & $67.84$ & $74.12$ & $70.95$\\
Lobby           & $68.42$ & $22.78$ & $20.35$ & $27.88$ & $21.64$\\
Trees           & $55.37$ & $64.01$ & $75.40$ & $85.80$ & $82.61$\\
WaterSurface    & $63.52$ & $51.16$ & $63.66$ & $78.16$ & $75.80$\\
\hline
Average         & $59.02$ & $53.34$ & $53.71$ & $67.36$ & $67.21$\\
%\hline\hline
%Average without Lobby    & $53.18$ & $52.28$ & $57.75$ & $XX.XX$ & %$63.70$ &$XX.XX$ & $XX.XX$ &$$\\
\hline
\end{tabular}
\end{center}
\caption{\e{F-measure} comparison between various existing algorithms on I2R data. Modeling the spatial influence of pixels (jKDE and DFB) significantly 
    improves accuracy.
    MoG and ACMMM03 results are as reported by Li \etal~\cite{Li10}. 
    For KDE, jKDE, and DFB, we use color dimension covariance value of 
    $45/4$ for both the background and foreground models. 
    For jKDE and DFB, we use spatial dimension covariance values of 
    $3/4$ and $12/4$ for the background and foreground models 
    respectively.} 
\label{tbl:early_methods}
\end{table}
%-------------------------------------------------------------------------
%-------------------------------------------------------------------------
\section{Adaptive kernel variances for the background likelihood}\label{sec:adaptive}
In this section we discuss recent improvements to our KDE likelihood model. 
Although KDE is a non-parametric approach to estimate probability 
densities, the choice of the kernel variance or the bandwidth 
is an important one. 
Using large bandwidth values can result in a very smooth
density function while low bandwidth values result in insufficient 
smoothing of the density function. 

\begin{figure}[t]
\begin{center}
   \includegraphics[width=1.0\linewidth]{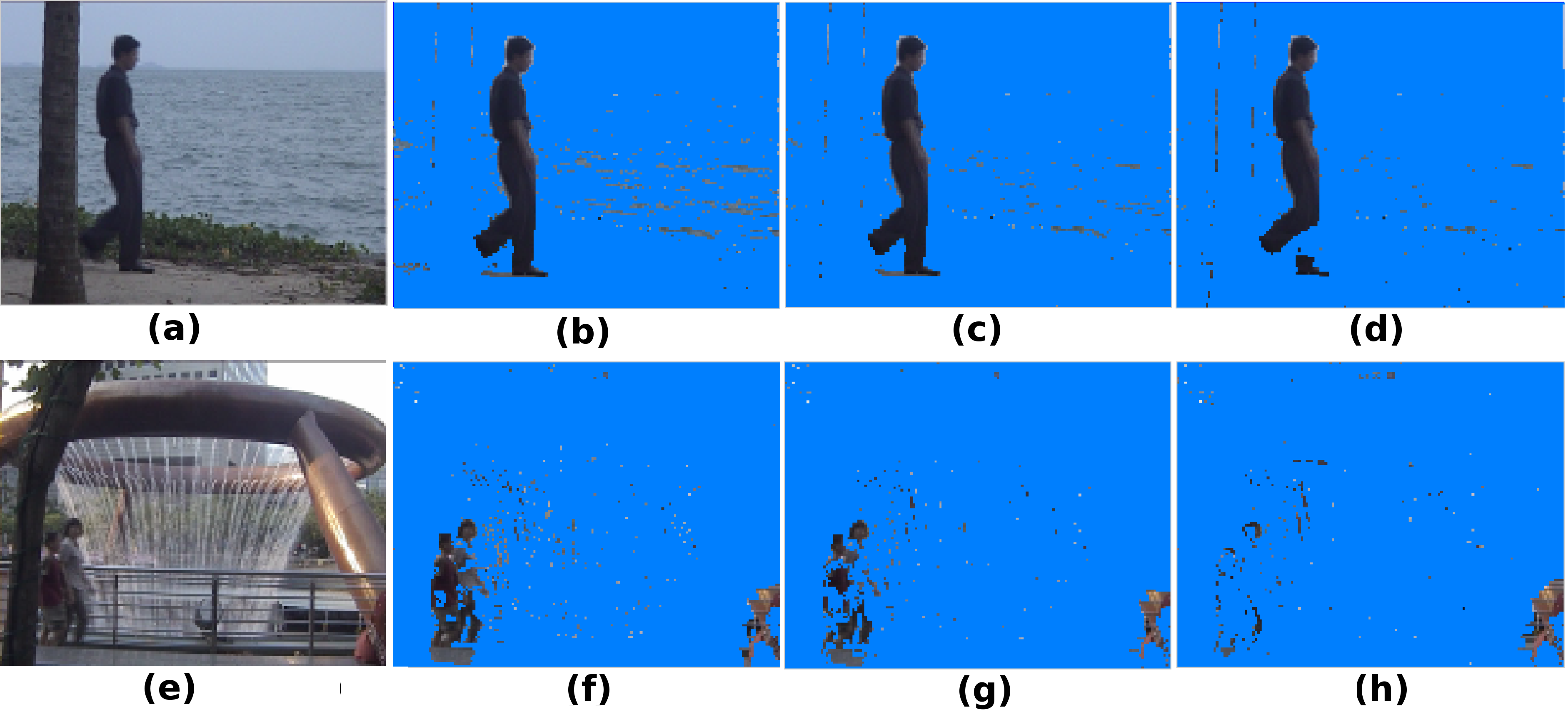}
\end{center}
   \caption{Two video sequences classified using increasing values of spatial kernel
       variance. Column 1 - original image, Column 2 - low variance,
                Column 3 - medium variance, Column 4 - high variance}
\label{fig:sigma_ex}
\end{figure}

In the context of background modeling, different parts of a dynamic scene 
may exhibit different statistics over 
the feature values and hence may need to be explained by different kernel 
variance values. 
Consider the result from a slightly different KDE 
model~\cite{Narayana12} shown in Figure \ref{fig:sigma_ex}. 
The figure shows background classification
results when the background likelihoods were
computed with increasing values of spatial dimension variance for two
different videos. 
Recall from Section \ref{sec:bgLik} that the spatial
variance controls the amount of influence that neighboring pixels have on
a given pixel's background likelihood.
Figures \ref{fig:sigma_ex}a to \ref{fig:sigma_ex}d show that 
having a high spatial dimension kernel variance helps in accurate 
classification of the water surface pixels, but doing so causes some 
pixels on the person's leg to become part of the background. 
Ideally, we would have different kernel variances
for the water surface pixels and the rest of the pixels. Similarly in the
second video (Figure \ref{fig:sigma_ex}e to \ref{fig:sigma_ex}h), having a high
kernel variance causes incorrect classification of many foreground pixels. 

Kernel variance selection for KDE is a well studied 
problem~\cite{Turlach93}, which can be addressed with variable-sized 
kernels~\cite{Wand95}. The kernel size or variance can be adapted at the 
estimation point (\e{balloon estimator}) or at each data sample 
point (\e{sample-point estimator}). 
Zivkovic and Heijden~\cite{Zivkovic06} use a balloon estimator to adapt the
kernel variance.
Mittal and Paragios~\cite{Mittal04} use a hybrid approach
but require that the uncertainty in the features be known. 

Using a different parameter for each pixel location can be useful in 
accounting for the varied nature of the background phenomenon at each pixel.
For the MoG model, Zivkovic~\cite{Zivkovic04} describes a method to find the
optimal number of Gaussians to use at each pixel.
For KDE models, Tavakkoli \etal~\cite{Tavakkoli09} learn the variance
for each pixel from a training set of frames, but do not 
adapt the learned values during the classification stage.

To address this problem, in earlier work~\cite{Narayana12,Narayana12BMVC}, 
we proposed a location-specific variance and an adaptive method to select 
the best variance at each location.
For each pixel location, for the background model, a set of
variance values for both spatial and color dimensions is tried and the
configuration that results in the highest likelihood is chosen for that
particular pixel. 

\begin{figure}[ht]
\begin{center}
\includegraphics[width=4in]{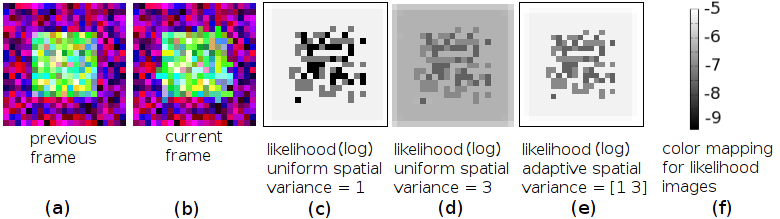}
\end{center}
   \caption{(a) and (b) Spatial uncertainty in the central part of the
    background. (c) Small uniform variance results in low likelihoods 
    for pixels that have moved. 
    (d) Large uniform variance results in higher likelihoods of
    the moved pixels at the expense of lowering the likelihoods of
    stationary pixels. (e) Adaptive variance results in high
           likelihoods for both the moved and stationary pixels.}
\label{fig:sharpen_xy}
\end{figure}

\begin{figure}[ht]
\begin{center}
\includegraphics[width=4in]{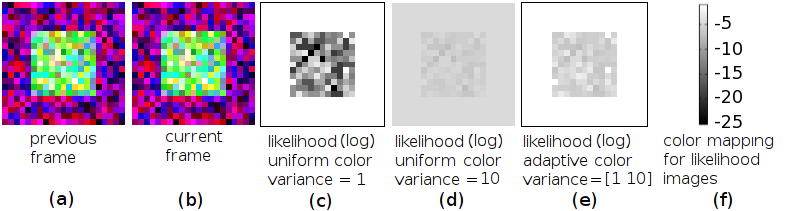}
\end{center}
   \caption{Color uncertainty in the central part of the background is
       best modeled by using adaptive kernel variances. (c) Small uniform
           variance results in low likelihoods for pixels that have
           changed color. (d) Large uniform variance results in higher likelihoods of
           the altered pixels at the expense of lowering the likelihoods of
           other pixels. (e) Adaptive variance results in high
           likelihoods for both kinds of pixels.}
\label{fig:sharpen_rgb}
\end{figure}

The effect of the adaptive kernel variance method can be interpreted easily
in Figures~\ref{fig:sharpen_xy} and \ref{fig:sharpen_rgb} (figures are 
originally from~\cite{Narayana12BMVC}).
Consider a synthetic scene with no foreground objects, but in which the colors in
the central
greenish part of the background have been displaced at random by one or two
pixel locations to simulate spatial uncertainty. As shown in
Figure~\ref{fig:sharpen_xy}, the adaptive kernel variance method models 
the scene better by applying a high spatial variance for pixels that have
moved and a low spatial variance for pixels that have not moved.
Similarly, for color variance, Figure~\ref{fig:sharpen_rgb} shows the resulting
likelihoods when uniformly sampled noise is added to the color
values in the central part of the image. A small color variance value
results in low likelihoods for pixels whose colors have changed. A large
color variance results in low likelihoods for pixels that have not changed.
The adaptive kernel variance method performs well in both kinds of pixels.

This improved background likelihood can be plugged into our system without
any changes to the rest of the system. The following section discusses
the increased accuracy that results from the substitution.
%-------------------------------------------------------------------------
%-------------------------------------------------------------------------
\section{Comparison}\label{sec:results}
Table \ref{tbl:all_methods} shows the results after using the adaptive 
kernel variance likelihood for the background.
We compare our system to a very successful background system that 
uses recently developed complex texture features called scale invariant
local ternary patterns (SILTP)~\cite{Li10} in a MoG model. 
These features are 
specifically designed to be robust to lighting changes and soft shadows
in the scene and represent the state of the art accuracy on this benchmark.
Results from the joint domain-range model with the use
of the adaptive variance likelihood (abbreviated as jKDE-A) show a 
decrease in accuracy compared to the earlier likelihood (jKDE). 
Using the adaptive procedure in our system (DFB-A) results in a remarkable 
increase in accuracy. Using simple color features, our system is able to
achieve accuracy comparable to SILTP on many videos. 

Using a combination of color and texture features has been shown to be
useful for background modeling~\cite{Yao07,Narayana12}. Texture features
are robust to lighting changes but not effective on large texture-less
objects. Color features are effective on large objects, but not very 
robust to varying illumination.
Including the SILTP 
feature representation along with LAB color features, which are more
robust to lighting changes, and performing
background modeling in this hybrid color-texture space returns the 
best results on a majority of videos. The parameters used for the 
adaptive kernel variance method and explanations of the 
improvement in the results are detailed in our earlier work~\cite{Narayana12BMVC}.

Our results are poor on two videos in the set - escalator and lobby.
The escalator video is from an indoor mall scene with a fast moving 
escalator. 
The escalator pixels exhibit a large amount of motion causing
them to be incorrectly classified as foreground in many frames.
The lobby video is from an office scene where a light switch is 
turned on and off at various times during the video.
Our likelihood model fails during the light switching and our use of 
an explicit foreground model causes the background model to take a very 
long time to recover.
Use of LAB color features and SILTP features helps in the drastic
illumination change scenario of the lobby video.
\subsection*{\textbf{Processing times}}
Our unoptimized matlab code for distribution field background modeling with 
adaptive variance for each pixel (DFB-A) takes 10 seconds per frame for 
videos of size 128 x 160 pixels.
In comparison, our implementation of the Sheikh and Shah model and our 
DFB model without the adaptive variance selection takes 5 seconds per frame.
In earlier work~\cite{Narayana12}, we describe a scheme to reduce 
computation time with the adaptive kernel method by recording the best 
variance values for each pixel from the previous frame. 
These cached variance values are first
used to classify pixels in the current frame. 
The expensive variance adaptation is performed only for
pixels where a confident classification is not achieved using the cached 
variance values. 
The caching method reduces the processing time to about 6 seconds per frame.
\begin{table}[ht]
\begin{center}
\begin{tabular}{|l|c|c|c|c|c|c|}
\hline
Video   & SILTP~\cite{Li10} & jKDE & jKDE-A & DFB & DFB-A & DFB-A\\ 
features& siltp             & rgb  & rgb    & rgb & rgb   & lab+siltp\\ 
\hline\hline
AirportHall     & $68.02$ & $70.13$ & $65.52$ & $67.95$ &$68.28$ &$\tb{70.75}$\\
Bootstrap       & $72.90$ & $71.77$ & $71.38$ &$69.17$ &$71.86$ &$\tb{77.64}$\\
Curtain         & $92.40$ & $87.34$ & $79.76$ & $85.66$ &$93.57$ &$\tb{94.07}$\\
Escalator       & $\tb{68.66}$ & $53.70$ & $54.02$ & $54.01$ &$66.37$ &$49.99$\\
Fountain        & $85.04$ & $57.35$ & $49.89$ &$77.11$ &$77.43$ &$\tb{85.88}$\\
ShoppingMall    & $79.65$ & $74.12$ & $74.43$ &$70.95$ &$76.46$ &$\tb{82.64}$\\
Lobby           & $\tb{79.21}$ & $27.88$ & $33.34$ &$21.64$ &$13.24$ &$62.60$\\
Trees           & $67.83$ & $85.80$ & $85.57$ &$82.61$ &$83.88$ &$\tb{87.64}$\\
WaterSurface    & $83.15$ & $78.16$ & $64.03$ &$75.80$ &$\tb{93.81}$ &$93.79$\\
%\hline\hline
%Average         & $63.11$ & $XX.XX$ & $73.16$ & $76.70$ &$76.40$ & $80.14$ &$74.85$ &$$\\
%Average without Lobby    & $53.18$ & $52.28$ & $57.75$ & $XX.XX$ & %$63.70$ &$XX.XX$ & $XX.XX$ &$$\\
\hline
\end{tabular}
\end{center}
\caption{\e{F-measure} on I2R data. Using the adaptive kernel variance method 
    with LAB color features and SILTP texture features results in the highest
        accuracy. Compared to uniform kernel variance DFB model, the 
        adaptive variance method DFB-A is more accurate.}
\label{tbl:all_methods}
\end{table}
%-------------------------------------------------------------------------
%-------------------------------------------------------------------------
\section{Discussion}\label{sec:discussion}
We argue that the view of background modeling described in this paper 
is, from a probabilistic perspective, clean and complete for the purpose of
background modeling.
By separating the various aspects of a background modeling system, namely 
the background likelihood, the foreground likelihood, and a prior, into
distinct components, we have presented a simple view of background 
modeling. For inference, these separate components are brought together 
to compute the posterior probability for background.

Previous backgrounding systems have also modeled the components that we 
have described, but have often combined them with each other or caused 
dependence between the components and the inference. 
The separation of the components from each other and their isolation from 
the inference step makes the system easy to understand and extend.
The individual components can be improved without having to consider 
their interdependence and effect on the inference.
We have shown one example of improving the background likelihood model and 
its positive impact on the system's accuracy.

We use a spatially varying prior that depends on 
the labels from the previous frame. The model can further
be improved by using a different 
prior at the image boundaries where new foreground objects are more 
likely. The modeling of the prior can also be improved by the explicit 
use of object tracking information.

We also believe that isolation of the model components can help in 
the development of effective learning methods for each of them. 
For example, the prior can be learned simply by counting the 
number of times each pixel is labeled as background or foreground. 
Maintaining a record of the number of times a pixel changes its label
from background to foreground and vice-versa is one
possible scheme to learn the prior values described in 
Section~\ref{sec:prior}. Such a learning scheme can help build a dynamic
model for the priors at different regions in the image.
%-------------------------------------------------------------------------
%-------------------------------------------------------------------------
\begin{acknowledgements}
This work was supported in part by the National Science Foundation under CAREER award
IIS-0546666 and grant CNS-0619337. Any opinions, findings, conclusions, or 
recommendations expressed here are the authors' and do not necessarily reflect 
those of the sponsors.
\end{acknowledgements}

\bibliographystyle{spmpsci}      % mathematics and physical sciences
\bibliography{dis_fields_backgrounding_bib.bib}

\end{document}